\documentclass[letterpaper, 10 pt, conference]{ieeeconf}  % Comment this line out if you need a4paper

\IEEEoverridecommandlockouts                              % This command is only needed if 
\usepackage{amssymb}  % assumes amsmath package installed
\usepackage{multirow}
\usepackage{algorithm, mathtools}
\usepackage{algpseudocode}
\usepackage{diagbox}
\usepackage[table,xcdraw]{xcolor}

\title{\LARGE \bf
Improving Sleep Scoring Generalization Through Meta Self-Supervision}
\title{\LARGE \bf
Towards Sleep Scoring Generalization\\ Through Self-Supervised Meta-Learning}

\author{Abdelhak Lemkhenter$^{1}$ and Paolo Favaro$^{1}$% <-this % stops a space
\thanks{$^{1}$Department of Computer Science, University of Bern, Bern, Switzerland
        {\tt\small abdelhak.lemkhenter,paolo.favaro@inf.unibe.ch}}%
}

\DeclareMathOperator*{\argmin}{argmin}
\newcommand{\ie}{\emph{i.e.}}
 
\usepackage{fancyhdr}
\begin{document}

% \

\maketitle

\thispagestyle{fancy}
\pagestyle{empty}

%%%%%%%%%%%%%%%%%%%%%%%%%%%%%%%%%%%%%%%%%%%%%%%%%%%%%%%%%%%%%%%%%%%%
\begin{abstract}
% In this work we introduce a novel meta-learning method for sleep scoring based on self-supervised learning. Our approach aims at building models for sleep scoring that can generalize across different patients and recording sessions, but do not require a further adaptation step to the target data. Towards this goal, we build our method on top of the Model Agnostic Meta-Learning (MAML) framework. In our analysis, we show that MAML can be significantly boosted in performance by incorporating a self-supervised learning (SSL) stage. This SSL stage is based on a general purpose pseudo-task that limits the overfitting to the specific variability of the training dataset. We show that our proposed method outperforms the baseline methods and state of the art meta-learning methods on the Sleep Cassette, Sleep Telemetry, ISRUC, UCD and CAP datasets.

In this work we introduce a novel meta-learning method for sleep scoring based on self-supervised learning. Our approach aims at building models for sleep scoring that can generalize across different patients and recording facilities, but do not require a further adaptation step to the target data. Towards this goal, we build our method on top of the Model Agnostic Meta-Learning (MAML) framework by incorporating a self-supervised learning (SSL) stage, and call it S2MAML. We show that S2MAML can significantly outperform MAML. The gain in performance comes from the SSL stage, which we base on a general purpose pseudo-task that limits the overfitting to the subject-specific patterns present in the training dataset. We show that S2MAML outperforms standard supervised learning and MAML on the SC, ST, ISRUC, UCD and CAP datasets.
\newline
\indent \textit{Clinical relevance}— Our work tackles the generalization problem of automatic sleep scoring models. This is one of the main hurdles that limits the adoption of such models for clinical and research sleep studies.
\end{abstract}

%%%%%%%%%%%%%%%%%%%%%%%%%%%%%%%%%%%%%%%%%%%%%%%%%%%%%%%%%%%%%%%%%%%%%%%%%%%%%%%%
\section{INTRODUCTION}
Sleep is known to play a significant role in the mental and physical health of an individual \cite{siegel2005clues} and therefore the development of tools to diagnose the quality of sleep and common sleep pathologies is fundamental. 
Sleep is monitored through polysomnography (PSG), \ie, the analysis of electrical bio-signals, such as the electroencephalogram (EEG), the electromyograph (EMG), the electrooculograph (EOG), and the electrocardiograph (ECG). The recorded bio-signals are split into 30-second intervals (epochs), and annotated by clinicians into several categories, such as: wake (W), non-rapid eye movement (NREM: N1, N2, and N3), and rapid eye movement (REM). This task is difficult and time-consuming. Thus, the ability to carry it out consistently and at a large scale through automation has an important impact on medical research and clinical practice \cite{wulff2010sleep}.

Towards this purpose, machine learning methods have been introduced as a way to obtain automatic sleep scoring \cite{fiorillo2021deepsleepnet, phan2019seqsleepnet,kemp2000sleepedf}. However, these methods are still not widely adopted amongst sleep practitioners. One limitation is that the current methods for sleep stage classification typically experience a decay in performance on data obtained from new cohorts of patients. 
The main reason behind this decay is the large variability of the bio-signals across subjects and sessions. This variability stems from experimental factors such as differences in the recording equipment and protocol (e.g., the number and placement of the electrodes) or physiological factors such as age, prognosis and medication. 
While this problem is commonly known among practitioners through direct experience, we illustrate it quantitatively in detail in our experimental analysis.

An approach to overcome this limitation is transfer learning, where a classification model is adapted to the target cohort through further training. This approach also motivated the recent work MetaSleepLearner \cite{banluesombatkul2020metasleeplearner}, which builds on meta-learning in the case of few-shot adaptation.   MetaSleepLearner requires only a small set of annotations of the target dataset and a limited amount of training whenever new data becomes available.

However, as we show in our experiments, the scheme used in MetaSleepLearner, \ie, MAML, still runs the risk of overfitting even if the adaptation is done on a large dataset. Moreover, we find the few-shot learning scenario or, more in general, the transfer learning case, not practical, because practitioners would need further training and/or to provide annotation to adapt a classifier to new target data. 
Thus, in this paper we propose to build a single sleep staging model and then use it ``as is'' on new data. To avoid the overfitting issues of the MetaSleepLearner we combine the Model Agnostic Meta-Learning (MAML) framework with self-supervised learning (SSL) \cite{lemkhenter2020boosting}. SSL has the advantage of not requiring annotation and it can be designed to train models that overfit less to the training data. With a slight abuse of notation we refer to the proposed setting as \emph{zero-shot learning}, to emphasize that no new training or annotation is needed with new data. 
To the best of our knowledge, the zero-shot learning scenario has not been explored so far in the literature for sleep scoring. 
We test our proposed method on several datasets and find that the use of SSL with MAML yields state of the art performance in zero-shot learning. 

\section{RELATED WORK}

The application of deep learning together with the steady increase in available public sleep data have resulted in a dramatic improvement of the performance of methods for automated sleep staging. These methods have now reached high levels of accuracy and robustness. Moreover, thanks to their computational efficiency they can be employed at a large scale and they can work directly on the raw data, instead of requiring hand-crafted guessing of useful pre-processing procedures. 

\subsection{Automatic Sleep Scoring}

Recent methods, such as SeqSleepNet \cite{phan2019seqsleepnet}, have focused on exploiting the context of the data by staging sequences rather than single epochs, or aimed at reducing the model parameters and introducing the estimation of the uncertainty of the prediction \cite{fiorillo2021deepsleepnet}.
One important limitation that has emerged is the lack of generalization, \ie, the drop in performance when trained models are used on new data. As mentioned in the Introduction, this phenomenon is currently attributed to the large diversity of the data across subjects/patients and sessions.
To address this problem, U-Sleep \cite{perslev2021u} introduces a u-net architecture for high frequency sleep staging. However, the generalization across datasets still remains an open problem.

In our method, we do not seek for an optimal architecture, but rather for a training scheme that can be easily adapted to other methods. Thus, for simplicity, we adopt the DeepSleepNet-Lite \cite{fiorillo2021deepsleepnet} as the classification model.

\subsection{Meta Learning}

A more fundamental approach towards generalization is the meta-learning framework \cite{finn2017model}. The main objective of meta-learning is to \emph{learn to learn}. In other words, rather than just learning from a single dataset to generalize to new data from the same distribution (as in the training set), one aims to train a model that can generalize well across other datasets.
A recent approach in this domain is Model Agnostic Meta Learning (MAML) \cite{finn2017model}. This method has been the employed successfully in recent work on sleep staging 
\cite{banluesombatkul2020metasleeplearner}, brain-computer interfacing \cite{li2021model} and for emotion prediction \cite{miyamoto2021meta}.
In our method, we combine it with self-supervised learning and show experimentally that this yields a significant boost in performance.

\subsection{Self-Supervised Learning}

Self-supervised learning (SSL) is a relatively recent technique in machine learning that has emerged as a very promising and powerful unsupervised learning approach. SSL allows one to train a model on data without annotation by specifying an artificial task, also called \emph{pseudo-task}. A model trained with such a pseudo-task can then be adapted to some target data through transfer learning. Recent work has defined pseudo-tasks based on detecting the phase-swap in EEG \cite{lemkhenter2020boosting} and the context prediction and contrastive predictive coding \cite{banville2021uncovering}. 
In our method, we propose to combine the ability to generalize of an SSL method \cite{lemkhenter2020boosting}
with the meta-learning framework.

\section{METHODS}
\subsection{Datasets}

In this work, we use 5 different sleep scoring datasets.

\subsubsection{Sleep Cassette (SC)} It is a subset of the Expanded Sleep-EDF Database \cite{kemp2000sleepedf} . It contains PSG sleep recordings obtained between 1981 and 1991. It includes recordings from 78 healthy subjects between the age of 25 and 101 with two recordings per person for most of them. 

\subsubsection{Sleep Telemetry (ST)} It is another subset of the Expanded Sleep-EDF Database \cite{kemp2000sleepedf}. It was collected as part of a 1994 study of the effect of temazepam on sleep. It contains PSG recordings from 22 subjects with one session per individual. 

Old datasets like SC \& ST allows us to investigate generalization from/to recordings with different signal quality.

\subsubsection{ISRUC} It is a publicly available sleep dataset \cite{khalighi2016isruc}. It consists of PSG recordings obtained at the Sleep Medicine Centre of the Hospital of Coimbra University (CHUC) between 2009 and 2013. This database has three different subsets:
\begin{itemize}
    \item Subgroup-I contains one recording per subject for 100 individuals with sleep disorders;
    \item Subgroup-II contains two recordings per subject for 8 individuals with sleep disorders; 
    \item Subgroup-III contains one recording per subject for 10 healthy individuals.
\end{itemize}

\subsubsection{University College Dublin Sleep Apnea Database (UCD)} It is a 2011 database collected at St. Vincent's University Hospital \cite{heneghan2011ucd}. It contains one PSG recording per subject for 25 individuals with suspected sleep-disordered breathing.

\subsubsection{Cyclic Alternating Pattern (CAP) Sleep Database} It is a collection of one recording per subject for 108 individuals with varying conditions. It contains 10 healthy subjects, 40 diagnosed with NFLE, 22 affected by RBD, 10 with PLM, 9 insomniac, 5 narcoleptic, 4 affected by SDB and 2 by bruxism \cite{terzano2001cap}. It was published in 2001.

\subsection{Data Preprocessing}

Out of all the signals available in each recording, we keep the EEG , EMG and EOG channels. All signals are re-sampled at 102.4Hz. This allows us to represent a 30sec epoch with 3072 time points, which is more compact and closer to the original sampling frequency compared to the commonly adopted 128Hz. This is sufficient since most spectral features classically used for sleep scoring are at lower frequency bands.

Since the convolution architecture we adopted in our experiments requires a constant number of channels as input, we fix that number to 9. If the recording contains more than 9 channels, which is the case for ISRUC and CAP, we randomly select a subset of them. Otherwise, if the recording does not have enough channels, we add dummy ones that are all zeros. The channels are shuffled before being fed to the model. We normalize each channel to have zero mean and a unit standard deviation.

\subsection{Data Split}
\label{sec:seenunseen}

To evaluate our models, we choose two different train/evaluation splits. We first split each dataset by \textbf{subjects}, then we randomly split each recording into samples of $3 \times 30$sec. This allows us to have an evaluation set containing subjects that were \textbf{seen} during training and another evaluation set containing \textbf{unseen} ones. Both splits follow a 75\%-25\% ratio. An illustration of both splits is shown in Table~\ref{tab:split}.

\begin{table}[t]
\centering
\caption{Diagram illustrating our two evaluation sets in a setting with 4 subjects and 4 samples per subject.}
\label{tab:split}
\begin{tabular}{|l|l|l|l|l|}
\hline
                   & \textbf{Sample 1} & \textbf{Sample 2} & \textbf{Sample 3} & \textbf{Sample 4} \\ \hline
\textbf{Subject 1} & Train             & Eval. Seen         & Train             & Train             \\ \hline
\textbf{Subject 2} & Train             & Train             & Train             & Eval. Seen         \\ \hline
\textbf{Subject 3} & Train             & Train             & Train             & Eval. Seen         \\ \hline
\textbf{Subject 4} & -                 & -                 & -                 & Val. Unseen       \\ \hline
\end{tabular}
\end{table}

\subsection{Notation}

We define the mapping $E: x \mapsto h$ as the encoding of the input signal $x$ into a feature vector $h$. The associated trainable parameters are denoted by $\Theta^E$. We denote the mapping from the feature vector $h$ to the predicted class label $\hat{y}$ for the supervised and self-supervised settings as $C^{SL}$ and $C^{SSL}$ respectively. Their associated trainable parameters are $\Theta_{SL}^C$ and $\Theta_{SSL}^C$ respectively. The predicted labels of the model are therefore \begin{equation}\label{eq:y_hat}
    \hat{y}^{SL/SSL} = C^{SL/SSL}(E(x)).
\end{equation}

In all experiments, models are trained by minimizing the average cross-entropy loss given by 
\begin{equation}\label{eq:ce}
    \mathcal{L}(T, \Theta^E, \Theta^C) = \frac{1}{|T|}\sum_{t \in T}  \frac{1}{|t|}\sum_{(x, y) \in t}- \sum_{c=1}^{N_c} y_c log(\hat{y}_c),
\end{equation}
where $y$, $\hat{y}$, $N_c$  and $T$ are respectively the true labels, the model predictions, the numbers of classes and a set of tasks $t$ consisting of signal-label pairs. Note that we represent the true labels as a one-hot encoding vector so $v_c$ refers to the $c$-th entry in the vector $v \in \mathbb{R}^{N_c}$. We frame the sleep scoring problem as a 5-way classification with the 5 classes being: Wake (W), N1, N2, N3 and REM.

\subsection{Self-Supervised MAML (S2MAML)}

Model Agnostic Meta Learning (MAML) is a meta-learning algorithm, where a given model is trained on a large variety of tasks with the goal of generalizing to novel tasks through fast-adaptation, \ie, few-shot learning or with no adaptation, \ie, zero-shot learning. In this work, we investigate the benefit of using meta-learning jointly with self-supervised learning to improve generalization to unseen subjects and datasets.

The problem that our proposed model solves can be described using the following bilevel formulation
% \begin{align}
%     &\Theta^{*E}, \Theta^{*C} = \argmin_{\Theta^{E}, \Theta^{C}_{SL}} \E_{t \sim T_{SL}} (\mathcal{L}( \{t\}, \hat{\Theta}^{E}, \Theta^{C}_{SL}))\\
%     &\text{s.t. } \hat{\Theta}^{E}, \Theta^{C}_{SSL} =  \argmin_{\Theta^{E}, \Theta^{C}_{SSL}} \E_{t' \sim T_{SSL}} (\mathcal{L}( \{t'\}, \Theta^{E}, \Theta^{C}_{SSL})), \nonumber
% \end{align}
\begin{align}
    \Theta^{*E}, \Theta^{*C} &= \argmin_{\Theta^{E}, \Theta^{C}_{SL}} \mathcal{L}( T_{SL}, \hat{\Theta}^{E}, \Theta^{C}_{SL})\\
    \text{s.t. } &\hat{\Theta}^{E}, \hat{\Theta}^{C}_{SSL} 
    =  \argmin_{\Theta^{E}, \Theta^{C}_{SSL}} \mathcal{L}( T_{SSL}, \Theta^{E}, \Theta^{C}_{SSL}), \nonumber
\end{align}
where the model $E$ is optimized to learn useful self-supervised representations of the set of tasks $T_{SSL}$. The bilevel optimization favors representations that generalize well to the supervised outer problem on tasks $T_{SL}$.

Algorithm~\ref{alg:main} outlines our adaptation of MAML, which we call \textbf{S2MAML}. Given $K$ datasets $\{D_k\}_{k=1}^K$, we randomly sample $n_{tasks}$ tasks from each one of them. Each task $t = \{(x_j, y_j)\}_{j=1}^{N_s}$  is defined as a set signal and label pairs belonging to the same subject in a given dataset. The total set of tasks $T$ is then split into a meta-training set $T^{tr}$ and a meta-validation set $T^{val}$. Each MAML iteration consists of an inner and an outer optimization problem. In the inner problem, $\Theta_{in}^{E}$ is initialized with the values of $\Theta^E$. Both $\Theta_{in}^{E}$ and $\Theta^{C}_{SSL}$ are optimized for $n_{in}$ iterations with respect to the self-supervised loss $\mathcal{L}^{in}$ computed on the meta-training set. For that, we need to generate a set of self-supervised tasks $T^{SSL}$ based on the $T^{tr}$. The details of this step are outlined in section~\ref{sec:ps}.

The weights $\Theta^E$ are then updated in the outer problem by minimizing the supervised loss $\mathcal{L}^{out}$ (see Algorithm~\ref{alg:main}) computed on the meta-validation set. The gradient for the $\Theta^E$ update is computed at $\Theta_{in}^E$, and not at $\Theta^E$, because we use the first-order approximation version of MAML \cite{finn2017model}. 

The goal of this design is to encourage the model to learn general purpose self-supervised features in the inner problem that would generalize well to the outer supervised problem computed on novel tasks, \ie, unseen subjects.

\begin{algorithm}[h]
\caption{S2MAML}\label{alg:main}
\begin{algorithmic}
\Require $\{D_k\}_{k=1}^K, \Theta^E, \Theta_{SSL}^C, \Theta_{Sup}^C, \lambda_{in}, \lambda_{out}$
\While{not converged}
    \State $T \gets \{\}$
    \For{ $k$ in $1..K$}
        \For{$i$ in $1..n_{tasks}$}
            \State $\mathrlap{t}\hphantom{T} \gets $ sample\_task($D_k$)
            \State $T \gets T \cup  \{t\}$
        \EndFor
    \EndFor
    
    \State $T^{tr}, T^{val} \gets $ split($T$)
    \State $\Theta^E_{in} \gets \Theta^E$
    
    \For{$i$ in $1..n_{in}$}
        \State $T_{SSL}^{tr} \gets$ generate\_ssl\_task($T^{tr}$)
        \State $\mathrlap{\mathcal{L}^{in}}\hphantom{T_{SSL}^{tr}}  \gets \mathcal{L}(T_{SSL}^{tr}, \Theta^E_{in}, \Theta_{SSL}^C)$
        \State $\mathrlap{\Theta^E_{in}}\hphantom{T_{SSL}^{tr}}  \gets \Theta^E_{in}  - \lambda_{in}\nabla_{\Theta^E_{in}}\mathcal{L}^{in}$
        \State $\mathrlap{\Theta_{SSL}^C}\hphantom{T_{SSL}^{tr}}  \gets \Theta_{SSL}^C - \lambda_{in}\nabla_{\Theta_{SSL}^C}\mathcal{L}^{in} $
    \EndFor
    
    \State $\mathrlap{\mathcal{L}^{out}}\hphantom{T_{SSL}^{tr}} \gets \mathcal{L}(T^{val}, \Theta^E_{in}, \Theta_{Sup}^C)$
    \State $\mathrlap{\Theta^E}\hphantom{T_{SSL}^{tr}} \gets \Theta^E  - \lambda_{out}\nabla_{\Theta^E_{in}}\mathcal{L}^{out}$
    \State $\mathrlap{\Theta_{Sup}^C}\hphantom{T_{SSL}^{tr}} \gets \Theta_{Sup}^C - \lambda_{out}\nabla_{\Theta_{Sup}^C}\mathcal{L}^{out} $
\EndWhile
\end{algorithmic}
\end{algorithm}

\subsection{PhaseSwap}\label{sec:ps}
For our self-supervised training, we choose \emph{PhaseSwap} (PS) introduced in \cite{lemkhenter2020boosting}. Our choice is motivated by two reasons. Firstly, PS has been shown to improve generalization to unseen subjects and this makes it a strong candidate for our approach. Secondly, PS can be defined on the same time scale as the supervised task. Indeed, other self-supervised methods such Relative Positioning (RP) or Contrastive Positional Coding (CPC) \cite{banville2021uncovering} require a longer temporal context, which would complicate the training loop. Since the main focus of this work is to highlight the potential of using self-supervised learning in a meta-learning setting, we opted for the simplest self-supervised loss. %to limit the number of additional hyper-parameters.

More specifically, PS is defined as a binary classification problem, where a model is trained to distinguish between samples $x$ and $x_{PS}$ defined as 
\begin{equation}\label{eq:ps}
    x_{PS} = \mathcal{F}^{-1} \left[ \left|\mathcal{F}\left(x\right)\right| \odot \measuredangle \mathcal{F}\left(x'\right)\right],
\end{equation}
where $x$ and $x'$ are two different samples. For a complex scalar $z \in \mathbf{C}^*$, the absolute value $|.|$ and angle $\measuredangle$ operators are defined such that $z = |z| e^{i\measuredangle z}$. 

In Algorithm~\ref{alg:main}, $T^{tr} = \{t_i\}_{i=1}^{K \times n_{tasks} / 2}$ is a set of supervised tasks. For each task $t \in T^{tr}$,  the function \mbox{\bf generate\_ssl\_task} generates a new task $t_{SSL}$ to be included in $T^{tr}_{SSL}$. For each signal-label pair $(x, y) \in t$, $t_{SSL}$ includes $(x, y_{SSL}=0)$ and its phase-swapped counterpart $(x_{PS}, y_{SSL}=1)$.  

\subsection{Architecture Choice}
For our experiment, we use DeepSleepNet-Lite \cite{fiorillo2021deepsleepnet} as our architecture of choice. It consists of two parallel convolutional neural networks using sets of small and large filters for the first layer respectively. The output of the two networks is concatenated into a single vector $h$ and fed into a softmax layer that maps it to the predicted class. The input $x$ to the network is a $90$sec segment, \ie, 3 consecutive epochs of $30$sec each.  We chose this architecture for its simplicity, its shorter temporal context and the fact that it does not require the power spectrum as input.

\subsection{Baselines and Training Hyper-parameters}

In all experiments, we compare the  performance of our S2MAML model to two other baselines: A supervised classification model without meta-learning and a MAML based training similar to ours, but where we replace the self-supervised problem in the inner loop with a supervised one. We refer to these models as \textbf{SL} and \textbf{MAML} respectively.

Unless stated otherwise, each task $t$ contains 8 samples from the same subjects. $n_{tasks}$, $n_{in}$, $\lambda_{out}$  and $\lambda_{in}$ are set to $32$, $1$, $10^{-4}$ and $5\cdot 10^{-5}$ respectively and each model is trained for 20 full iterations over all the databases considered for training. We use Adam \cite{kingma2015adam} as optimizer with its default hyperparameters. Our models are implemented using Pytorch\footnote{https://pytorch.org/} and ran on a single NVIDIA 1080Ti GPU. We observe no significant differences between the computation times off all models both in inference and training.

We adopt the same label smoothing regularization used by~\cite{fiorillo2021deepsleepnet} with their suggested tuning.

\subsection{Evaluation Metrics}

We use macro F1 (MF1) as an evaluation metric for our experiments. Macro F1 is defined as 

\begin{equation}\label{eq:macro}
    \text{MF1} = \frac{1}{N_c} \sum_{c=1}^{N_c} \text{F1}_c = \frac{1}{N_c} \sum_{c=1}^{N_c} \frac{2 \text{P}_c \times \text{R}_c}{\text{P}_c +  \text{R}_c}
\end{equation}

where $N_c$, $P_c$ and $R_c$ are respectively the number of classes, the precision and recall for the class $c$. It is the average F1 score per class, where the F1 score is defined as the harmonic mean of precision and recall. We choose MF1, instead of the classic F1 score, as it is a better metric when the data has significant class imbalance, which is the case for sleep scoring. All reported MF1 Scores are averaged across a 4-way cross validation split.
%%%%%%%%%%%%%%%%%%%%%%%%%%%%%%%%%%%%%%%%%%%%%%%%%%%%%%%%%%%%%%%%%%%%%%%%%%%%%%%%
\section{RESULTS}

\subsection{Generalization to Novel Databases: \textbf{3 vs 5}}\label{sec:3vs5}

\begin{table*}[h!]
\caption{Cross-validation MF1 Scores for the 3 vs 5 setting. Avg(S) and Avg(U) refer to the average MF1s across all seen and unseen evaluation sets marked with (S) and (U) respectively. Avg(U1) and Avg(U2) refer to the averages MF1s for unseen subjects across seen (CAP(U), ISRUC(U), ST(U)) and unseen (UCD(U), SC(U)) datasets respectively. Avg(U) is the average MF1 across all unseen sets.}
\label{tab:3vs5}
\centering

\setlength{\tabcolsep}{5pt}
\begin{tabular}{|lcccc|cccc|ccc|c|}
\hline
\textbf{Run} & \textbf{CAP(S)} & \textbf{ST(S)} & \textbf{ISRUC(S)} & \textbf{Avg(S)} & \textbf{CAP(U)} & \textbf{ST(U)} & \textbf{ISRUC(U)} & \textbf{Avg(U1)} & \textbf{SC(U)} & \textbf{UCD(U)} & \textbf{Avg(U2)} & \textbf{Avg(U)} \\ \hline
S2MAML         & \textbf{68.8}   & 74.8            & \textbf{74.7}     & \textbf{72.8}   & \textbf{56.5}   & 65.2            & \textbf{70.3}     & \textbf{64.0}    & \textbf{41.1}  & \textbf{43.7}   & \textbf{42.4}    & \textbf{55.4}   \\
MAML         & 66.4            & 71.3            & 73.3              & 70.3            & 55.0            & \textbf{65.3}   & 68.9              & 63.1             & 34.4           & 42.1            & 38.2             & 53.1            \\
SL         & 55.0            & \textbf{75.5}   & 66.7              & 65.7            & 46.4            & 63.2            & 63.0              & 57.5             & 30.3           & 37.9            & 34.1             & 48.1            \\ \hline
\end{tabular}
\end{table*}

In this set of experiments,  we compare the performance of S2MAML to the two baselines when training on 3 out of 5 of the considered databases. This allows us to evaluate the performance of our model on completely unseen cohorts of subjects belonging to different databases (see section~\ref{sec:seenunseen}). More specifically, we train using ST, CAP and ISRUC and evaluate on all five datasets. We report the performance of all models on both evaluation sets with \textbf{seen} and \textbf{unseen} subjects in Table~\ref{tab:3vs5}. 

For seen subjects, we observe that our model outperforms the two baselines (supervised and MAML training) on average as well as on CAP and ISRUC. The performance gap on average is of $2.5\%$ compared to the MAML and $7.1\%$ compared to supervised training. This shows that meta-learning based methods are generally better suited for overcoming intra-subject variability, and that self-supervision is a powerful tool to further reduce that performance gap.  

% \begin{table}[t]
% \centering
% \caption{REWRITE: Comparison of models trained on CAP, ST and ISRUC on evaluation set of subjects present in the training set. 4-way cv average}
% \label{tab:3vs5-seen}
% \begin{tabular}{lcccc}
% \hline
% \textbf{Run} & \textbf{CAP}  & \textbf{ST}   & \textbf{ISRUC} & \textbf{Avg}  \\ \hline
% S2MAML         & \textbf{68.8} & \textbf{74.8} & \textbf{74.7}  & \textbf{72.8} \\
% MAML         & 66.4          & 71.3          & 73.3           & 70.3          \\
% SL         & 55.0          & 75.5          & 66.7           & 65.7          \\ \hline
% \end{tabular}
% \end{table}

 For unseen subjects, we observe that our model outperforms both baselines on most datasets as well as on average. We also find that the meta-learning based models outperform the supervised based baseline, a feature that we noticed also with seen subjects. More importantly, the performance gap between our S2MAML and MAML is wider on held out datasets. Although MAML generalizes better to unseen subjects from the databases used for training compared to the supervised baseline, it generalizes less to held out databases compared to our model. In other words, our S2MAML is not only better suited for dealing with inter-subject variability, but it is also better suited for dealing with inter-cohort variability.

We discuss the low performance on ST in in section~\ref{sec:onevsall}.

% \begin{table}[h]
% \centering
% \caption{Reported MF1 for models trained on CAP, ST and ISRUC for unseen subjects during training. We also report the average of MF1s across the datasets used for training, the average of MF1s across novel datasets as well as the average across all datasets. 4-way CV average.}
% \label{tab:3vs5-unseen}
% \begin{tabular}{lcccc|ccc|c}
% \hline
% \textbf{Run} & \textbf{CAP}  & \textbf{ST}   & \textbf{ISRUC} & \textbf{Avg}  & \textbf{SC}   & \textbf{UCD}  & \textbf{Avg}  & \textbf{Avg}  \\ \hline
% Ous          & \textbf{56.5} & 65.2          & \textbf{70.3}  & \textbf{64.0} & \textbf{41.1} & \textbf{43.7} & \textbf{42.4} & \textbf{55.4} \\
% MAML         & 55.0          & \textbf{65.3} & 68.9           & 63.1          & 34.4          & 42.1          & 38.2          & 53.1          \\
% SL         & 46.4          & 63.2          & 63.0           & 57.5          & 30.3          & 37.9          & 34.1          & 48.1          \\ \hline
% \end{tabular}
% \end{table}

\subsection{Generalization in a Data Abundant Setting: \textbf{All vs All}}\label{sec:allvsall}

In this set of experiments we compare the performance of S2MAML to our two baselines when trained on all databases jointly. This allows us to highlight the benefit of our algorithm in a setting where a large quantity of labeled recordings are available. The MF1 Scores of all models on both seen and unseen subjects are reported in Table~\ref{tab:allvsall}.

\begin{table}[h]
\caption{Cross-validation MF1 Scores for the All vs All setting. We also report the average MF1 across all databases.}
\label{tab:allvsall}

\centering
\setlength{\tabcolsep}{5pt}
\begin{tabular}{|lccccccc|}
\hline
\textbf{Run} & \textbf{Subjects} & \textbf{CAP}  & \textbf{ST}   & \textbf{ISRUC} & \textbf{SC}   & \textbf{UCD}  & \textbf{Avg}  \\ \hline
S2MAML       & Seen              & \textbf{82.1} & \textbf{85.0} & \textbf{88.8}  & \textbf{86.3} & \textbf{90.4} & \textbf{86.5} \\
MAML         & Seen              & 80.2          & 81.5          & 86.1           & 84.1          & 89.4          & 84.3          \\
SL           & Seen              & 59.3          & 83.2          & 71.2           & 82.1          & 68.6          & 72.9          \\ \hline
S2MAML       & Unseen            & \textbf{67.9} & \textbf{73.7} & \textbf{82.7}  & \textbf{83.8} & \textbf{70.7} & \textbf{75.8} \\
MAML         & Unseen            & 65.7          & 69.2          & 80.8           & 80.8          & 69.8          & 73.3          \\
SL           & Unseen            & 50.8          & 70.0          & 67.2           & 79.5          & 55.0          & 64.5          \\ \hline
\end{tabular}
\end{table}

We observe that our model outperforms both baselines on all datasets as well as on average for both evaluation settings. This shows that the generalization advantage of our model does not disappear when scaling up the amount of available data. In the deep learning literature, scaling up the amount of training data is a common practice used to improve the generalization of artificial neural networks. This relies on the implicit assumption that with enough data, one is able to obtain a training set that is similar in distribution to the evaluation set and contains most sources of variability that can be encountered. However, in the case of physiological signals this assumption may not hold as well. A new individual will always have subject-specific sleep patterns and the inter-dataset variability will always remain a challenge as long as hardware/software recording pipelines keep evolving.

In the previous section, we have split the data by subjects, which is not a common practice. We did so to illustrate a more extreme setting for generalization. In this section, we also obtain the performance on seen subjects, as done in the literature, so that it is easier to compare to prior work.  Although the main focus of our work is to reduce the generalization gap between subjects and datasets, the MF1s reported on seen subjects are comparable or better than state of the art methods in the literature. Our model achieves an MF1 score of $86.3\%$ and $85.0\%$ compared to $79\%$ and $76\%$ for U-Sleep \cite{perslev2021u} on SC and ST respectively. The numbers are however not directly comparable due to the difference in the randomness of the splits. For this reason, and in order to keep our results focused on the generalization problem, we chose to omit numbers reported by other prior works from our tables.

\subsection{Disparity Between Datasets: \textbf{One vs All}}\label{sec:onevsall}
In this set of experiments, we compare the performance of the different models on unseen databases when trained only on a single one. This represents a worst case scenario, where one has access to a very limited number of subjects and therefore learning to generalize becomes much more challenging. Since the different databases considered in this study have different sizes, we choose to equalize experiments by training for a fixed number of gradient updates, $5000$, instead of looping through the training set $20$ times. The goal of these experiments is to gauge how similar or dissimilar the databases considered in this work are. In other words, our goal is to confirm that generalizing from one set to the others is indeed a challenging task and that each database has its particularities. 

We report the obtained MF1 scores on Table~\ref{tab:onevsall}. For all datasets and all three models considered, we observe that the performance drops significantly on unseen datasets. One additional noteworthy observation is that out of all combinations, models trained on SC/ST and tested on others and vice versa seem to generalize the least. On the other hand, generalizing between ST and SC seems more feasible. This may be due to fact that ST/SC were collected a few decades ago or the fact that they include EEG electrodes that are not common in the other three databases. We believe that this observation may explain why both meta-learning models struggle compared to the SL baseline on ST as reported in Table~\ref{tab:3vs5}.

Overall in this setting the performance across different methods does not indicate a clear winner. Given the restricted number of subjects per dataset, all methods struggle to learn features that generalize well to new cohorts. However, on average across all possible combinations, S2MAML and MAML comes out slightly on top with $29.8 \%$ and $29.9 \%$ respectively compared to $29.4 \%$ for the SL baseline. 

\begin{table}[t]
\caption{Cross-validation MF1 Scores for models trained on one dataset and evaluated on unseen subjects/datasets. Each row block corresponds to models trained on a single dataset.}
\label{tab:onevsall}
\centering
\begin{tabular}{|llccccc|}
\hline
\diagbox[width=6em]{\textbf{Train}}{\textbf{Test}} & \textbf{Model} & \textbf{ISRUC}                        & \textbf{SC}                           & \textbf{ST}                           & \textbf{CAP}                          & \textbf{UCD}                          \\ \hline
                        & S2MAML & \cellcolor[HTML]{34A853}\textbf{76.0} & \cellcolor[HTML]{EAD465}27.4          & \cellcolor[HTML]{F07244}9.7           & \cellcolor[HTML]{F0D665}\textbf{25.8} & \cellcolor[HTML]{F3864B}12.2          \\
                        & MAML   & \cellcolor[HTML]{3AAA54}74.7          & \cellcolor[HTML]{F5D766}\textbf{24.5} & \cellcolor[HTML]{F17746}\textbf{10.3} & \cellcolor[HTML]{FAD866}23.3          & \cellcolor[HTML]{F69E53}15.0          \\
\multirow{-3}{*}{ISRUC} & SL     & \cellcolor[HTML]{45AC55}71.7          & \cellcolor[HTML]{FAB65A}17.7          & \cellcolor[HTML]{F17746}10.3          & \cellcolor[HTML]{FED765}21.6          & \cellcolor[HTML]{F8A856}\textbf{16.1} \\ \hline
                        & S2MAML & \cellcolor[HTML]{ED5C3D}7.2           & \cellcolor[HTML]{3CAA54}74.1          & \cellcolor[HTML]{B6C860}41.5          & \cellcolor[HTML]{EC543A}6.3           & \cellcolor[HTML]{ED5E3D}7.4           \\
                        & MAML   & \cellcolor[HTML]{EE643F}8.1           & \cellcolor[HTML]{38A954}75.2          & \cellcolor[HTML]{ABC55F}\textbf{44.3} & \cellcolor[HTML]{FCC45F}\textbf{19.4} & \cellcolor[HTML]{ED5F3E}7.5           \\
\multirow{-3}{*}{SC}    & SL     & \cellcolor[HTML]{F17545}\textbf{10.1} & \cellcolor[HTML]{37A954}\textbf{75.5} & \cellcolor[HTML]{ADC65F}43.7          & \cellcolor[HTML]{EC573B}6.6           & \cellcolor[HTML]{EE6540}\textbf{8.2}  \\ \hline
                        & S2MAML & \cellcolor[HTML]{F69E52}14.9          & \cellcolor[HTML]{E6D364}28.7          & \cellcolor[HTML]{54B056}67.7          & \cellcolor[HTML]{F9B259}17.2          & \cellcolor[HTML]{FDCB61}\textbf{20.2} \\
                        & MAML   & \cellcolor[HTML]{FCC35F}19.3          & \cellcolor[HTML]{E5D364}\textbf{28.9} & \cellcolor[HTML]{52B056}68.2          & \cellcolor[HTML]{FABC5C}\textbf{18.4} & \cellcolor[HTML]{FBC25E}19.2          \\
\multirow{-3}{*}{ST}    & SL     & \cellcolor[HTML]{FED865}\textbf{21.7} & \cellcolor[HTML]{E7D464}28.3          & \cellcolor[HTML]{51AF56}\textbf{68.3} & \cellcolor[HTML]{FABA5C}18.2          & \cellcolor[HTML]{FCC660}19.6          \\ \hline
                        & S2MAML & \cellcolor[HTML]{E5D364}28.8          & \cellcolor[HTML]{EAD465}\textbf{27.4} & \cellcolor[HTML]{F59550}13.9          & \cellcolor[HTML]{78B95A}\textbf{58.1} & \cellcolor[HTML]{9CC15D}\textbf{48.5} \\
                        & MAML   & \cellcolor[HTML]{E6D364}28.5          & \cellcolor[HTML]{F5D765}24.6          & \cellcolor[HTML]{F48F4E}13.2          & \cellcolor[HTML]{86BC5B}54.3          & \cellcolor[HTML]{A3C35E}46.4          \\
\multirow{-3}{*}{CAP}   & SL     & \cellcolor[HTML]{E2D264}\textbf{29.8} & \cellcolor[HTML]{F2D665}25.4          & \cellcolor[HTML]{F69A51}\textbf{14.4} & \cellcolor[HTML]{9AC15D}49.0          & \cellcolor[HTML]{B5C760}41.7          \\ \hline
                        & S2MAML & \cellcolor[HTML]{FAD866}23.2          & \cellcolor[HTML]{FBBD5D}18.6          & \cellcolor[HTML]{EA4535}4.5           & \cellcolor[HTML]{F6D766}24.2          & \cellcolor[HTML]{68B558}\textbf{62.2} \\
                        & MAML   & \cellcolor[HTML]{FCCA61}20.0          & \cellcolor[HTML]{FBBD5C}\textbf{18.5} & \cellcolor[HTML]{EA4836}4.9           & \cellcolor[HTML]{FDD163}20.9          & \cellcolor[HTML]{73B859}59.4          \\
\multirow{-3}{*}{UCD}   & SL     & \cellcolor[HTML]{E1D264}29.9          & \cellcolor[HTML]{FBBD5C}18.5          & \cellcolor[HTML]{EC533A}\textbf{6.1}  & \cellcolor[HTML]{F1D665}\textbf{25.7} & \cellcolor[HTML]{79B95A}57.7          \\ \hline
\end{tabular}
\end{table}

\subsection{Effect of $\lambda_{in}$}

In this section we study the effect of $\lambda_{in}$ on our model and the MAML baseline. We train both our model and the MAML baseline in the \textbf{3 vs 5} setting described in Section~\ref{sec:3vs5} for $\lambda_{in} \in \{10^{-3}, 5\cdot 10^{-5}\}$. Tables~\ref{tab:lr-seen}
and \ref{tab:lr-unseen} report the obtained MF1 scores for seen and unseen subjects respectively. We observe that while the value of $\lambda_{in}$ has little effect on the performance of our model, setting it to $10^{-3}$ greatly reduces the performance of the MAML baseline. By setting $\lambda_{in}$ to a higher value, we put more emphasis on the convergence on the meta-train set, \ie, in the inner loop. This confirms that using \emph{PhaseSwap} as a self-supervised task in the inner loop, \ie, on the meta-train set, is less prone to learning subject-specific features and thus generalizes better compared to its supervised counterpart. Additionally, this shows that our methods is more robust to the choice of the hyper-parameter $\lambda_{in}$.

\begin{table}[t]
\centering
\caption{Cross-validation MF1 Scores on unseen subjects for models trained in the 3 vs 5 setting for different values of $\lambda_{in}$.}
\label{tab:lr-unseen}
\setlength{\tabcolsep}{4pt}
\begin{tabular}{|lccccc|ccc|}
\hline
\textbf{Run} & $\lambda_{in}$   & \textbf{CAP}  & \textbf{ST}   & \textbf{ISRUC} & \textbf{Avg}  & \textbf{SC}   & \textbf{UCD}  & \textbf{Avg}  \\ \hline
S2MAML       & $10^{-3}$        & 60.0          & \textbf{70.2} & 67.4           & 65.9          & \textbf{34.3} & \textbf{49.0} & \textbf{41.7} \\
S2MAML       & $5\cdot 10^{-5}$ & \textbf{61.9} & 70.1          & \textbf{68.4}  & \textbf{66.8} & 32.6          & 46.9          & 39.8          \\ \hline
MAML         & $10^{-3}$        & 23.1          & 25.4          & 33.1           & 27.2          & 15.9          & 10.4          & 13.2          \\
MAML         & $5\cdot 10^{-5}$ & \textbf{59.2} & \textbf{67.9} & \textbf{65.4}  & \textbf{64.2} & \textbf{25.9} & \textbf{49.4} & \textbf{37.7} \\ \hline
\end{tabular}
\end{table}

\begin{table}[t]
\centering
\caption{Cross-validation MF1 Scores on seen subjects for models trained in the 3 vs 5 setting for different values of $\lambda_{in}$.}
\label{tab:lr-seen}
\begin{tabular}{|llcccc|}
\hline
\textbf{Run} & $\lambda_{in}$   & \textbf{CAP}  & \textbf{ST}   & \textbf{ISRUC} & \textbf{Avg}  \\ \hline
S2MAML       & $10^{-3}$        & 68.3          & 70.8          & 73.4           & 71.0          \\
S2MAML       & $5\cdot 10^{-5}$ & \textbf{68.8} & \textbf{74.8} & \textbf{74.7}  & \textbf{72.8} \\ \hline
MAML         & $10^{-3}$        & 16.6          & 24.5          & 20.5           & 20.5          \\
MAML         & $5\cdot 10^{-5}$ & \textbf{66.4} & \textbf{71.3} & \textbf{73.2}  & \textbf{70.8} \\ \hline
\end{tabular}
\end{table}

%%%%%%%%%%%%%%%%%%%%%%%%%%%%%%%%%%%%%%%%%%%%%%%%%%%%%%%%%%%%%%%%%%%%%%%%%%%%%%%%
\section{DISCUSSIONS}

With the increasing popularity of deep learning methods, more and more artificial neural network architectures have been proposed in the literature for automatic sleep scoring. Reliable automatic sleep scoring models have the potential to speed up sleep research and make it more accessible by reducing the cost of manual annotations and enable more advanced closed-loop system. However, one important requirement for such models is that they should maintain their level of performance across sessions, subjects and hardware/software recording settings. Our work positions itself as a step forward toward achieving this goal. By leveraging both meta-learning and self-supervised learning, our S2MAML is able to reduce the performance drop associated with both intra-subject variability, \ie, unseen subjects from seen datasets, and intra-database variability, \ie, on unseen datasets. 

%%%%%%%%%%%%%%%%%%%%%%%%%%%%%%%%%%%%%%%%%%%%%%%%%%%%%%%%%%%%%%%%%%%%%%%%%%%%%%%%
\section{CONCLUSIONS}
In this work, we introduce a novel deep learning model for automatic sleep scoring. By leveraging meta-learning and robust self-supervised features, our model is able to better cope with intra-subject and intra-dataset variabilities in the zero-shot setting. We show through extensive experiments that our model outperforms all baselines in terms of generalization capabilities both on seen and unseen subjects. Our work presents itself as an important milestone toward the wide adoption of automatic sleep scoring in sleep research by bridging the performance gap present when deploying such models on new datasets and cohorts of subjects.

\section*{ACKNOWLEDGMENT}
This research was supported by the Interfaculty Research Cooperation ``Decoding Sleep: From Neurons to Health \& Mind'' of the University of Bern.
\bibliographystyle{ieeetr}
\bibliography{biblio}

\end{document}